\documentclass{article} 
\usepackage[preprint]{colm2026_conference}

\usepackage{microtype}
\usepackage{hyperref}
\usepackage{url}
\usepackage{booktabs}

\usepackage{lineno}

\definecolor{darkblue}{rgb}{0, 0, 0.5}
\hypersetup{colorlinks=true, citecolor=darkblue, linkcolor=darkblue, urlcolor=darkblue}

\usepackage{xcolor}
\usepackage{multirow}
\usepackage{multicol}
\usepackage{makecell}
\usepackage{amsmath}
\usepackage{amssymb}
\usepackage{graphicx}
\usepackage{subcaption}
\usepackage{wrapfig}

\newcommand{\revised}[1]{#1}
\newcommand{\deltaup}[1]{\textsubscript{\textcolor{green!70!black}{+#1}}} 
\newcommand{\deltaneg}[1]{\textsubscript{\textcolor{gray!90!white}{#1}}} 
\newcommand{\deltadown}[1]{\textsubscript{\textcolor{pink!90!black}{-#1}}} 
\newcommand{\deltabigdown}[1]{\textsubscript{\textcolor{red!80!black}{-#1}}} 

\title{Building Multi-Task Agentic LLMs via Two-Phase Distillation}

\author{Huaijie Wang$^1$\thanks{This work was done during the internship at Ant Group.} \quad Shusheng Xu$^2$ \quad Yi Wu$^1$ \quad Kaifeng Lyu$^1$ \\
${}^1$Tsinghua University \quad ${}^2$Ant Group \\
\small \texttt{wang-hx23@mails.tsinghua.edu.cn} \enspace \texttt{\{xssstory,jxwuyi\}@gmail.com} \enspace \texttt{klyu@mail.tsinghua.edu.cn}}

\begin{document}

\ifcolmsubmission
\linenumbers
\fi

\maketitle

\begin{abstract}
    A key step toward artificial general intelligence is to train models that can perform multiple tasks. In this paper, we study how to build such models by first training separate RL experts for individual tasks and then consolidating them via distillation, as an alternative to directly training a single model on mixed tasks. We show that off-policy distillation degrades in multi-task settings due to the mode-covering nature of forward KL: aggregating data from multiple tasks introduces a large number of behavioral modes that can exceed the student's capacity, forcing it to average across behaviors and leading to degraded performance. In contrast, on-policy distillation is mode-seeking but requires strong initialization.
    Inspired by these observations, we propose a two-phase approach: off-policy distillation followed by on-policy refinement.
    Evaluation across conversational agents and text-based games confirms that this two-phase approach matches single-task RL expert performance for each individual task, whereas off-policy or on-policy distillation alone fails to match this performance.
\end{abstract}

\section{Introduction}

Reinforcement Learning (RL) has demonstrated remarkable success across a wide range of agentic large language model (LLM) tasks, including interactive tool-using agents~\citep{gao2026selfevolvingsyntheticdataverifiablereward}, search agents~\citep{gao2025turnsunlockinglonghorizonagentic}, and text-based games~\citep{liu2025gemgymagenticllms}. As an important step towards artificial general intelligence, LLMs are expected to solve multiple tasks simultaneously. However, building such unified multi-task models remains challenging. A straightforward approach is multi-task RL, i.e., jointly training a single model on multiple tasks. Unfortunately, multi-task RL presents significant practical challenges. \citet{wu2025imbalancedgradientsrlposttraining} find that different tasks may exhibit vastly different gradient scales, yet larger gradients do not necessarily correlate with greater performance gains. \citet{ramesh2026multitaskgrporeliablellm} observe that certain tasks can dominate the training process, leading to suboptimal multi-task performance. Furthermore, multi-task RL may incur substantial computational costs, as the training cost of each trial is comparable to the sum of training all individual tasks.

An alternative solution is to train one expert model for each task and then combine these models into a unified model. Two primary approaches exist for such combination: parameter merging~\citep{ilharco2023editingmodelstaskarithmetic,yu2024languagemodelssupermario} and distillation~\citep{deepseekai2025deepseekv32pushingfrontieropen,coreteam2026mimov2flashtechnicalreport,agarwal2024onpolicydistillationlanguagemodels}. While parameter merging offers computational efficiency by requiring little or no additional training, it operates as a black box with limited interpretability. Parameter merging often suffers from performance trade-offs between tasks~\citep{wang2026mixmergemultidomainreinforcement}. Empirically, we observe substantial performance degradation with parameter merging on the benchmarks we evaluate. Additionally, the effectiveness of parameter merging is highly sensitive to the training dynamics of the supervised fine-tuning (SFT) phase. (See Appendix~\ref{sec:additional_results} for details.)

This paper focuses on distillation-based methods, which provide a more interpretable paradigm for combining knowledge from different expert models. We consider two main distillation approaches: \textit{off-policy} and \textit{on-policy} distillation.
Off-policy distillation collects rollouts from teacher models and applies supervised fine-tuning, essentially optimizing the forward KL divergence. In contrast, on-policy distillation samples from the student model and optimizes the reverse KL divergence.

\paragraph{Limitations of Off-Policy and On-Policy Distillation.}
Comparing off-policy and on-policy distillation, off-policy distillation is easier to implement and offers better training stability,
but prior work has identified several limitations of off-policy distillation: (1) exposure bias~\citep{bengio2015scheduledsamplingsequenceprediction,agarwal2024onpolicydistillationlanguagemodels,xu2025kdrlposttrainingreasoningllms,gu2026minillmonpolicydistillationlarge}, where the student model makes out-of-distribution errors during generation but receives no guidance on recovery from the training data, causing errors to accumulate across generation steps; (2) mode-covering behavior~\citep{agarwal2024onpolicydistillationlanguagemodels,gu2026minillmonpolicydistillationlarge}, which typically occurs when the teacher has larger capacity than the student—since off-policy distillation optimizes forward KL (a mode-covering objective), the student model is incentivized to place probability mass across all modes in the teacher distribution, including placing excessive probability on low-probability regions; (3) conflicting reasoning structures between teacher and student~\citep{chen2026molecularstructurethoughtmapping}.

Among these possible limitations of off-policy distillation, our experiments reveal that when building multi-task agentic LLMs, the major challenge likely stems from its mode-covering property. Even when the student and teacher models share the same base model, aggregating data from multiple tasks introduces a larger number of behavioral modes that can exceed the student model's capacity, forcing it to compromise by averaging across modes. Conversely, on-policy distillation cannot be applied directly without proper initialization, as it requires the student model to already possess sufficiently strong performance to generate reasonable trajectories.

\paragraph{Our Method: Two-Phase Distillation.}
To address this limitation, we propose a two-phase distillation approach: off-policy distillation followed by on-policy refinement. The mode-seeking property of on-policy distillation enables the student model to focus on a subset of modes, mitigating the averaging behavior induced by off-policy distillation. Meanwhile, off-policy distillation provides the necessary initialization for on-policy distillation to generate reasonable trajectories. We evaluate our method on four tasks: two conversational agent domains from $\tau^2$-bench~\citep{barres2025tau2benchevaluatingconversationalagents} and two text-based games from GEM~\citep{liu2025gemgymagenticllms}, demonstrating that our approach successfully recovers the performance of single-task RL models. Additionally, we investigate data aggregation strategies for multi-task off-policy distillation, showing that while data mixture affects off-policy distillation performance, our two-phase approach remains robust across different mixture configurations.
\section{Related Works}
\subsection{Knowledge Distillation}
Distillation transfers knowledge from a teacher model to a student model through two primary approaches. Off-policy distillation, also known as supervised knowledge distillation, collects data from the teacher model and applies supervised fine-tuning (SFT) to the student~\citep{hinton2015distillingknowledgeneuralnetwork}. Recent successes~\citep{guha2025openthoughtsdatarecipesreasoning,moshkov2025aimo2,du2025nemotronmath} have demonstrated that off-policy distillation enables small models (e.g., 1.5B parameters) to acquire strong reasoning capabilities for complex tasks in mathematics, coding, and science. However, as argued by \citet{agarwal2024onpolicydistillationlanguagemodels}, the mode-covering nature of forward KL divergence in off-policy distillation can cause the student to assign excessive probability mass to low-probability regions of the teacher's distribution, leading to hallucination and performance degradation. Another concern is exposure bias~\citep{bengio2015scheduledsamplingsequenceprediction}, where the student's sampling distribution at inference time may deviate from the teacher distribution, with errors potentially accumulating across generation steps. However, the severity of exposure bias in autoregressive language models remains debated~\citep{he2021exposurebiasversusselfrecovery}. An alternative paradigm, on-policy distillation~\citep{agarwal2024onpolicydistillationlanguagemodels,lu2025onpolicydistillation,gu2026minillmonpolicydistillationlarge}, samples from the student's own distribution rather than the teacher's. This approach has been successfully incorporated to build powerful small language models from larger ones~\citep{gu2026minillmonpolicydistillationlarge,yang2025qwen3technicalreport,bai2025qwen3vltechnicalreport}. However, on-policy distillation requires the student model to be sufficiently capable, as learning is limited to trajectories the student can already generate.

When unifying multiple domain-specific LLMs into a single model, both distillation approaches have been explored. \citet{deepseekai2025deepseekv32pushingfrontieropen} employ off-policy distillation and observe that while it introduces marginal performance drops, these can be recovered through a subsequent mixed RL phase. \citet{coreteam2026mimov2flashtechnicalreport} incorporate on-policy distillation and achieve performance comparable to domain experts on most benchmarks. Our experiments reveal that proper initialization is critical for on-policy distillation: relying solely on SFT for initialization can lead to unstable training and final performance that falls behind domain specialists.

\subsection{Model Merging}
Model merging operates directly in parameter space to combine multiple task-specific LLMs into a unified model. A straightforward approach fuses the parameter deltas of different models~\citep{ilharco2023editingmodelstaskarithmetic}. Subsequent works address parameter interference through various techniques, such as neglecting minority signs~\citep{yadav2023tiesmergingresolvinginterferencemerging,wan2024fusechatknowledgefusionchat,meituanlongcatteam2025introducinglongcatflashthinkingtechnicalreport} or applying dropout to parameter deltas~\citep{yu2024languagemodelssupermario,meituanlongcatteam2025introducinglongcatflashthinkingtechnicalreport}.
\citet{wang2026mixmergemultidomainreinforcement} introduce the concept of policy neighborhoods, where merging neighboring policies can improve performance. However, they demonstrate that this relationship is asymmetric and that merging non-neighboring policies may result in performance trade-offs, with gains in one task accompanied by degradation in another.
An alternative approach determines optimal weights for averaging parameter deltas using metrics such as the Fisher Information Matrix~\citep{matena2022mergingmodelsfisherweightedaveraging,lee2025dynamicfisherweightedmodelmerging} or task-specific evaluation metrics~\citep{lee2025dynamicfisherweightedmodelmerging,li2025maplowcomputemodelmerging}. However, these methods require additional computation and can be complex to implement.
\section{Preliminary}
We consider two types of distillation methods for combining single-task RL experts: off-policy distillation (Sec.~\ref{sec:off_policy}) and on-policy distillation (Sec.~\ref{sec:on_policy}).

\subsection{Off-Policy Distillation}\label{sec:off_policy}
Off-policy distillation transfers knowledge from expert teacher models to a student model by collecting rollouts generated by the teachers and training the student via supervised fine-tuning (SFT). Given a teacher model $\pi_T(y|x)$ and a training prompt dataset $\mathcal{D}_\text{train}$, we collect $K$ rollouts from each training prompt:
\begin{equation}
    \mathcal{D}_\text{rollout}=\bigcup_{x\in\mathcal{D}_\text{train}}\{(x,y_1),(x,y_2),\ldots,(x,y_K)\mid y_i\sim \pi_T(\cdot|x)\}.
\end{equation}
The student model $\pi_\theta(y|x)$ is then trained to minimize the negative log-likelihood:
\begin{equation}
    \mathcal{L}_\text{SFT} = \mathbb{E}_{(x,y)\sim\mathcal{D}_\text{rollout}}[-\log\pi_\theta(y|x)].
\end{equation}
This objective is equivalent to minimizing the forward KL divergence $\operatorname{KL}[\pi_T \| \pi_\theta]$ between the teacher and student distributions.

In the multi-task setting, we simultaneously distill from multiple expert models, each trained specifically for a different task. Let $\{\pi_T^{(i)}\}_{i=1}^N$ denote the set of $N$ task-specific expert models. For each expert, we collect rollouts from its corresponding task dataset and merge these rollout data to form a unified training set $\mathcal{D}_\text{multitask} = \bigcup_{i=1}^{N} \mathcal{D}_\text{rollout}^{(i)}$, where $\mathcal{D}_\text{rollout}^{(i)}$ contains the rollouts collected from expert $\pi_T^{(i)}$. The student model is then trained on this merged dataset using the SFT objective.

\subsection{On-Policy Distillation}\label{sec:on_policy}
On-policy distillation~\citep{agarwal2024onpolicydistillationlanguagemodels,lu2025onpolicydistillation,coreteam2026mimov2flashtechnicalreport} aims to minimize the reverse KL divergence $\operatorname{KL}[\pi_\theta \| \pi_T]$ between the student model and the teacher model. This requires sampling from the student model, which cannot be performed in advance. In this paper, we follow the methodology of \citet{lu2025onpolicydistillation,coreteam2026mimov2flashtechnicalreport}. The reverse KL divergence can be written as
\begin{align}
    &\quad\operatorname{KL}[\pi_\theta(\cdot|x) \| \pi_T(\cdot|x)] \\
    &= \mathbb E_{y\sim\pi_\theta(\cdot|x)}[\log\pi_\theta(y|x) - \log\pi_T(y|x)] \\
    &= \mathbb E_{y\sim\pi_\theta(\cdot|x)}\left[ \sum_{i=1}^L \left(\log\pi_\theta(y_i|xy_{1:i-1}) - \log\pi_T(y_i|xy_{1:i-1})\right) \right]. \label{eq:kl}
\end{align}
Leveraging this decomposition, we apply reinforcement learning with the reward given by
\begin{equation}
    \revised{r(xy_{1:i}) = -(\log\pi_\theta(y_i|xy_{1:i-1}) - \log\pi_T(y_i|xy_{1:i-1})).}
\end{equation}
\revised{In this case, the KL divergence in Eq.~\ref{eq:kl} is the negated accumulated reward $\mathbb E_{y\sim\pi_\theta(\cdot|x)}[\sum_{i=1}^L-r(xy_{1:i})].$}
We follow \citet{lu2025onpolicydistillation} and set the discount factor $\gamma=0$. In this case, we can directly adopt $\hat A_i=-(\log\pi_\theta(y_i|xy{1:i-1}) - \log\pi_T(y_i|xy_{1:i-1}))$ as the advantage estimation in the policy gradient objective.

Notably, since on-policy distillation leverages samples from the student model, it is expected that the student model already possesses adequate performance~\citep{agarwal2024onpolicydistillationlanguagemodels}.

In the multi-task setting, we extend on-policy distillation to simultaneously imitate multiple teacher models on their respective tasks. Specifically, each training batch contains prompts sampled from different task datasets. For a prompt from task $i$, we apply the on-policy distillation objective using the corresponding teacher model $\pi_T^{(i)}$ to compute the reward. This allows the student model to learn from all expert models during training.
\section{Main Results}
We begin by presenting results for existing distillation approaches: off-policy distillation and on-policy distillation applied in isolation. On-policy distillation is initialized from SFT. We also evaluate two alternative approaches (multi-task RL and parameter merging) in Appendix~\ref{sec:additional_results}.

\paragraph{Tasks and Evaluation Metrics.} We evaluate our methods on four diverse tasks spanning two domains. From $\tau^2$-bench~\citep{barres2025tau2benchevaluatingconversationalagents}, we select two conversational agent domains: \texttt{airline} (booking, modifying, and canceling flight reservations and handling refunds) and \texttt{telecom} (technical support, bill payment, line suspension, and plan options). These tasks demand multiple agent capabilities: calling domain-specific APIs following guidelines, interacting with user agents, and dual-control (guiding the user agent to correctly invoke APIs). We also incorporate two text-based games from GEM~\citep{liu2025gemgymagenticllms}: \texttt{sudoku} and \texttt{mastermind}, both using the \texttt{v0-hard} variant. These games require logical reasoning, planning, and memory recall over evolving game states~\citep{guertler2025textarena}. For \texttt{sudoku}, we extend the max number of turns to 60. We use success rate as the evaluation metric for all tasks. For the $\tau^2$-bench tasks, we sample 16 trajectories per test prompt and compute the average success rate. For the GEM text games, we randomly generate 128 intial setups and sample 16 trajectories for each game setup. For all 4 tasks, we report pass@1 (success rate per trajectory) and pass\^{}4 (probability that all 4 i.i.d. trials succeed). We perform experiments using two model scales, \texttt{Qwen3-8B} and \texttt{Qwen3-30B-A3B-Thinking-2507}, to evaluate the effectiveness of our approach across different model capacities.

For the $\tau^2$-bench tasks, we use a locally hosted \texttt{Qwen3-235B-A22B-Instruct-2507} model as the user simulator instead of \texttt{gpt-4.1-2025-04-14} to ensure more stable and reproducible results. For the \texttt{telecom} domain, the user model is further fine-tuned using rejection sampling to improve its tool calling ability.

\paragraph{Training Details.} For off-policy distillation, we sample rollouts from the teacher model using the RL training prompts. For the 8B models, we collect $128\times 400$ (prompt, answer) pairs per task, while for the 30B models, we use $128\times 800$ pairs per task to account for the larger model capacity. We train with a batch size of 128. For on-policy distillation, each batch contains 16 training prompts with 16 rollouts per prompt. 

\begin{table}[t]
    \centering
    \small
    \setlength{\tabcolsep}{4pt}
    \begin{tabular}{c|l|l|l|l|l|l|l|l}
        \toprule
        \multirow{3}{*}{Method} & \multicolumn{4}{c|}{$\tau^2$-bench} & \multicolumn{4}{c}{GEM} \\
         & \multicolumn{2}{c|}{airline} & \multicolumn{2}{c|}{telecom} & \multicolumn{2}{c|}{sudoku} & \multicolumn{2}{c}{mastermind} \\
         & \multicolumn{1}{c|}{pass@1} & \multicolumn{1}{c|}{pass\^{}4} & \multicolumn{1}{c|}{pass@1} & \multicolumn{1}{c|}{pass\^{}4} & \multicolumn{1}{c|}{pass@1} & \multicolumn{1}{c|}{pass\^{}4} & \multicolumn{1}{c|}{pass@1} & \multicolumn{1}{c}{pass\^{}4} \\
        \midrule
        \multicolumn{9}{c}{\textit{Single-Task}} \\
        \midrule
        Pre-RL & 42.5 & 26.5 & 66.2 & 44.7 & 2.2 & 0.0 & 3.6 & 0.0 \\
        RL & 53.5 & 33.6 & 85.9 & 67.8 & 92.8 & 73.7 & 77.3 & 38.7 \\
        Off-policy Distill. & 50.0\deltabigdown{3.5} & 32.1\deltadown{1.5} & 84.3\deltadown{1.6} & 67.7\deltaneg{-0.1} & 93.2\deltaneg{+0.4} & 75.4\deltaup{1.7} & 77.1\deltaneg{-0.2} & 41.7\deltaup{3.0} \\
        \midrule
        \multicolumn{9}{c}{\textit{Multi-Task}} \\
        \midrule
        Off-policy Distill. & 51.8\deltadown{1.7} & 31.3\deltadown{2.3} & 84.8\deltadown{1.1} & 64.9\deltabigdown{2.9} & 93.8\deltaup{1.0} & 77.5\deltaup{3.8} & 76.5\deltaneg{-0.8} & 39.6\deltaneg{+0.9} \\
        On-policy Distill. & 52.6\deltaneg{-0.9} & 32.9\deltaneg{-0.7} & 78.2\deltabigdown{7.7} & 58.3\deltabigdown{9.5} & 93.9\deltaup{1.1} & 77.8\deltaup{4.1} & 78.0\deltaneg{+0.7} & 41.8\deltaup{3.1} \\
        Off. + On. & 51.4\deltadown{2.1} & 33.2\deltaneg{-0.4} & 85.8\deltaneg{-0.1} & 69.4\deltaup{1.6} & 94.8\deltaup{2.0} & 81.0\deltaup{7.3} & 78.6\deltaup{1.3} & 43.1\deltaup{4.4} \\
        \bottomrule
    \end{tabular}
    \caption{Main results on \texttt{Qwen3-8B}. Colored subscripts show performance deltas versus single-task RL. Off-policy distillation recovers RL performance in single-task but degrades in multi-task (especially on \texttt{telecom}). Combining off-policy and on-policy distillation (``Off. + On.'') restores performance. On-policy alone fails without proper initialization.}
    \label{tab:main}
\end{table}

\begin{table}[t]
    \centering
    \small
    \setlength{\tabcolsep}{4pt}
    \begin{tabular}{c|l|l|l|l|l|l|l|l}
        \toprule
        \multirow{3}{*}{Method} & \multicolumn{4}{c|}{$\tau^2$-bench} & \multicolumn{4}{c}{GEM} \\
         & \multicolumn{2}{c|}{airline} & \multicolumn{2}{c|}{telecom} & \multicolumn{2}{c|}{sudoku} & \multicolumn{2}{c}{mastermind} \\
         & \multicolumn{1}{c|}{pass@1} & \multicolumn{1}{c|}{pass\^{}4} & \multicolumn{1}{c|}{pass@1} & \multicolumn{1}{c|}{pass\^{}4} & \multicolumn{1}{c|}{pass@1} & \multicolumn{1}{c|}{pass\^{}4} & \multicolumn{1}{c|}{pass@1} & \multicolumn{1}{c}{pass\^{}4} \\
        \midrule
        \multicolumn{9}{c}{\textit{Single-Task}} \\
        \midrule
        Pre-RL & 52.9 & 30.9 & 77.4 & 51.1 & 5.1 & 0.0 & 1.0 & 0.0 \\
        RL & 61.6 & 42.6 & 92.8 & 76.4 & 98.7 & 94.9 & 93.5 & 77.1 \\
        Off-policy Distill. & 60.5\deltadown{1.1} & 41.7\deltaneg{-0.9} & 92.0\deltaneg{-0.8} & 74.7\deltadown{1.7} & 98.9\deltaneg{+0.2} & 95.7\deltaneg{+0.8} & 94.6\deltaup{1.1} & 81.0\deltaup{3.9} \\
        \midrule
        \multicolumn{9}{c}{\textit{Multi-Task}} \\
        \midrule
        Off-policy Distill. & 61.0\deltaneg{-0.6} & 41.5\deltadown{1.1} & 88.0\deltabigdown{4.8} & 65.8\deltabigdown{10.6} & 98.2\deltaneg{-0.5} & 93.0\deltadown{1.9} & 91.5\deltadown{2.0} & 71.8\deltabigdown{5.3} \\
        On-policy Distill. & 61.1\deltaneg{-0.5} & 41.0\deltadown{1.6} & 88.8\deltabigdown{4.0} & 72.2\deltabigdown{4.2} & 99.1\deltaneg{+0.4} & 96.4\deltaup{1.5} & 93.4\deltaneg{-0.1} & 77.0\deltaneg{-0.1} \\
        Off. + On. & 62.0\deltaneg{+0.4} & 43.8\deltaup{1.2} & 92.2\deltaneg{-0.6} & 74.7\deltadown{1.7} & 99.3\deltaneg{+0.6} & 97.1\deltaup{2.2} & 93.9\deltaneg{+0.4} & 78.4\deltaup{1.3} \\
        \bottomrule
    \end{tabular}
    \caption{Main results on \texttt{Qwen3-30B-A3B-Thinking-2507}. Colored subscripts show performance deltas versus single-task RL. Results are consistent with the 8B model: off-policy distillation degrades in multi-task while ``Off. + On.'' recovers performance.}
    \label{tab:main_30b}
\end{table}
We present our main results in Tab.~\ref{tab:main} and Tab.~\ref{tab:main_30b}. We highlight 3 key findings below.

\begin{figure}[t]
    \centering
    \begin{subfigure}[b]{0.48\textwidth}
        \centering
        \includegraphics[width=\textwidth]{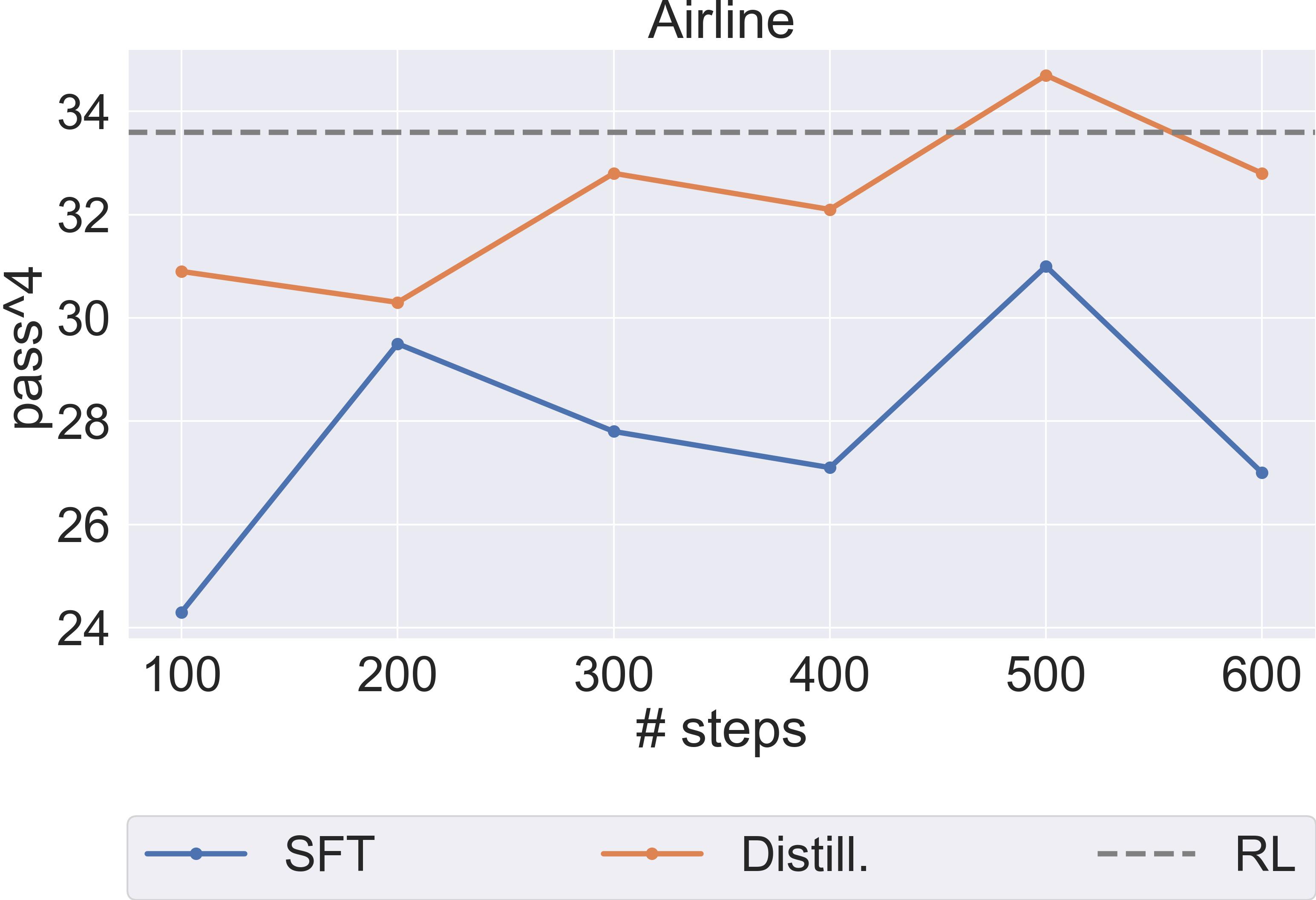}
    \end{subfigure}
    \hfill
    \begin{subfigure}[b]{0.48\textwidth}
        \centering
        \includegraphics[width=\textwidth]{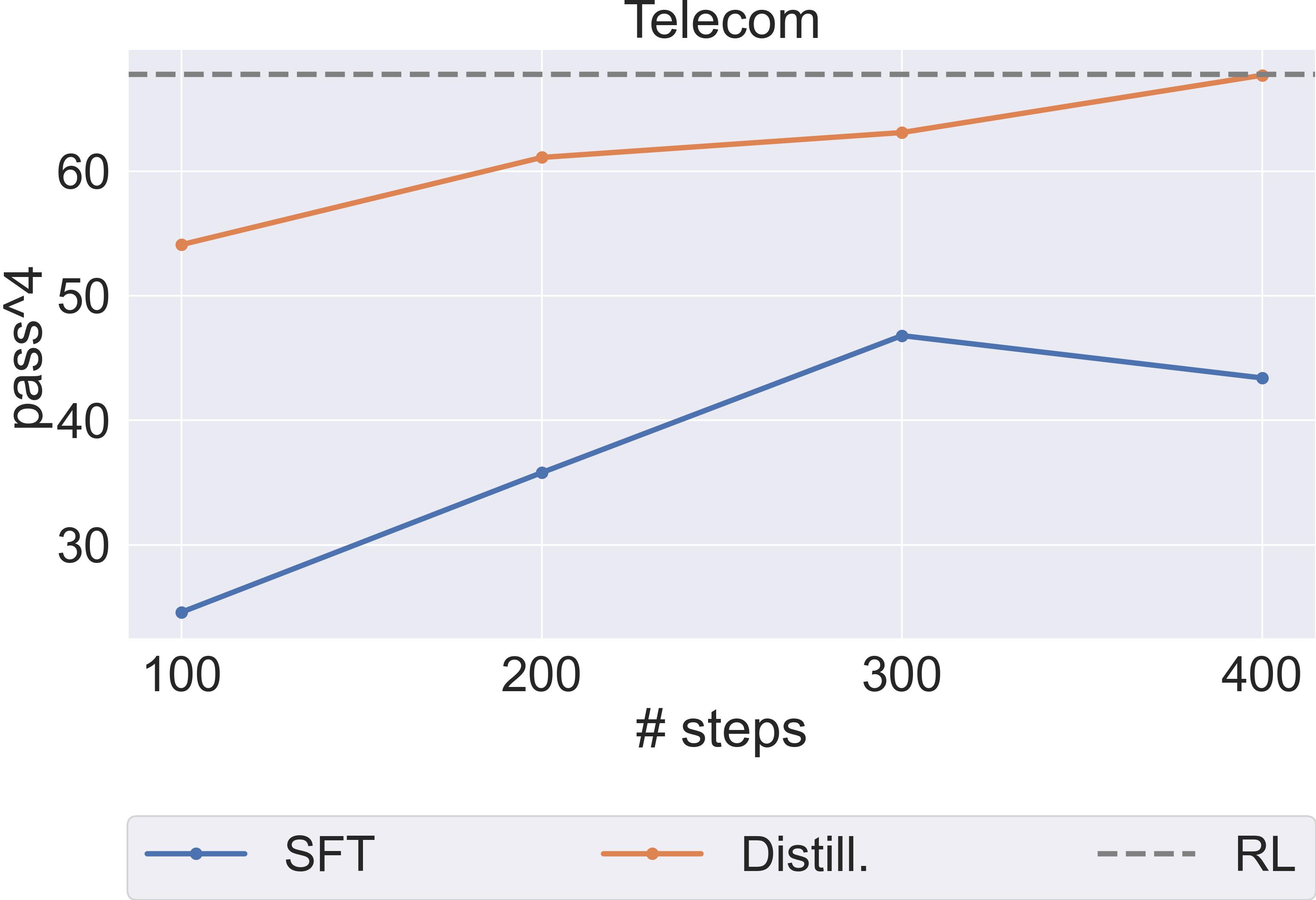}
    \end{subfigure}
    \caption{Learning curves comparing SFT and off-policy distillation (single-task setting) on \texttt{airline} and \texttt{telecom}, \revised{where off-policy distillation is essentially SFT on teacher-generated data.} Despite \revised{the SFT baseline} leveraging rollouts from a substantially stronger model, off-policy distillation from RL models trained on the same base model achieves superior performance under a limited training budget.}
    \label{fig:sft_vs_distill}
\end{figure}

\paragraph{Off-policy distillation recovers performance of single-task RL models.} In the single-task setting, off-policy distillation successfully recovers the performance of RL experts when applied to the same base model, demonstrating that expert trajectories can effectively transfer learned behaviors. Interetingly, the SFT phase before RL leverages data collected from \texttt{claude-opus-4.5}, so it is effectively performing off-policy distillation from a stronger model. However, as presented in Fig.~\ref{fig:sft_vs_distill}, under a modest training budget, distilling from the RL model (which is built upon the same base model) achieves even better performance than distilling from a much stronger model. Similar observations have been made in recent work~\citep{gu2026minillmonpolicydistillationlarge}, attributing this to teacher models containing more modes than students can capture.

\begin{wrapfigure}[16]{r}{0.3\textwidth}
    \centering
    \includegraphics[width=0.28\textwidth]{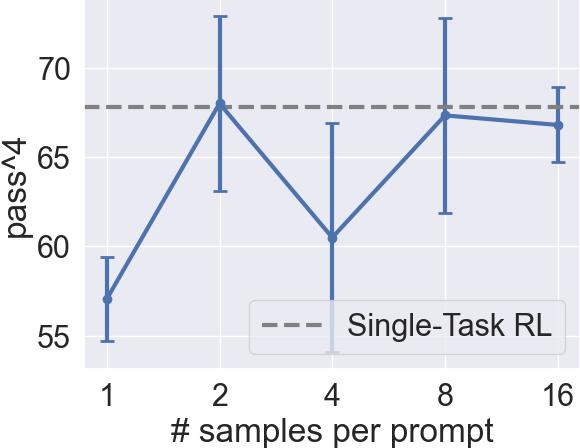}
    \caption{Off-policy distillation performance (pass\^{}4) on \texttt{telecom} with varying samples per prompt. Dataset size fixed at $128\times 400$. \revised{Performance degrades with fewer samples per prompt.}}
    \label{fig:nsamples}
\end{wrapfigure}

We also investigate the number of samples per prompt required for successful single-task off-policy distillation. For the \texttt{telecom} task, we vary the number of samples per prompt and then upsample (or downsample) to maintain a fixed dataset size of $128\times 400$. Notably, sampling 16 rollouts per training prompt produces 63,191 samples, exceeding $128\times 400$. We report \texttt{telecom} pass\^{}4 performance in Fig.~\ref{fig:nsamples}. \revised{We find that using fewer samples per prompt (and therefore upsampling the data) harms performance, resulting in poor results (1 sample per prompt) or high variance across trials (2-8 samples per prompt). In contrast, using 16 samples per prompt (only 1 epoch) yields much lower variance.}

\paragraph{Multi-task off-policy distillation exhibits performance degradation.} Despite its simplicity, off-policy distillation in the multi-task setting achieves performance comparable to single-task RL models on most tasks, even surpassing them on \texttt{sudoku}. However, we observe notable performance degradation on \texttt{telecom} (8B: -2.9\% pass\^{}4, 30B: -10.6\% pass\^{}4), suggesting that simple aggregation of expert trajectories may not always suffice for multi-task learning. Notably, the degradation is more severe for pass\^{}4 than pass@1, indicating that the model struggles to consistently produce the desired behavior. Since single-task off-policy distillation achieves comparable performance to RL, we rule out exposure bias as the primary cause. Instead, we hypothesize that the root cause is the mode-covering behavior of off-policy distillation. We provide more discussion of this phenomenon in Sec.~\ref{sec:offpolicy_vs_onpolicy}.

\paragraph{On-policy distillation may require proper initialization.} When applied directly without off-policy distillation in advance, on-policy distillation fails to achieve satisfactory performance. As shown in Tab.~\ref{tab:main_30b}, it exhibits substantial performance drops on \texttt{telecom} (-9.5\% pass\^{}4 on 8B models, -4.2\% pass\^{}4 on 30B models), with degradation particularly severe for the 8B model. One possible explanation is that SFT models may not provide sufficiently strong initialization for on-policy distillation. This aligns with previous findings that on-policy distillation requires the model to have sufficiently strong initial performance, as the student model must generate reasonable trajectories for effective learning~\citep{agarwal2024onpolicydistillationlanguagemodels}.

\section{Our Method}\label{sec:offpolicy_vs_onpolicy}
Given the limitations of both distillation methods, we address two key questions: Why does off-policy distillation fail in multi-task settings? How can we robustly distill multiple experts into a unified multi-task model? We attribute the failure of off-policy distillation to the mode-covering property of forward KL divergence. In Sec.~\ref{sec:toy}, we present a toy example illustrating why mode-covering could become problematic in multi-task settings. Building on this insight, we propose a two-phase distillation recipe and demonstrate that our method successfully recovers the performance of single-task RL baselines. Finally, in Sec.~\ref{sec:validate}, we further validate our hypothesis through experiments that manipulate the off-policy distillation data.

\subsection{A Toy Example}\label{sec:toy}
We present a toy example illustrating why off-policy distillation could fail in multi-task settings. Consider a task where the input prompt $p\in\{0,1\}$ requires generating two tokens $a_1\in\{0,1\}$ and $a_2\in\mathcal S$ ($\mathcal S\supset\{0,1\}$ is a sufficiently large set). The reward function is given by $\mathbb{I}\{p=a_2\}$, i.e., $a_2$ must match $p$, while $a_1$ represents intermediate reasoning. $p=0$ and $p=1$ can be viewed as two ``domains''. We consider a model $M$ with limited capacity: $a_1\sim p_1^M(\cdot|p)$ and $a_2\sim p_2^M(\cdot|a_1)$, where each token depends only on the previous token and position.

The optimal policy for domain $p=0$ has $p_1^M$ being arbitrary and $p_2^M$ always outputs 0; similarly for $p=1$ with $p_2^M$ outputting 1. Without loss of generality, we assume both optimal policies have uniform $p_1^M$ distributions. Collecting expert trajectories yields dataset $\mathcal{D}=\{000,010,101,111\}$. Consider two student policies: $U$ (uniform) where both distributions $p_1^M,p_2^M$ are uniform over $\{0,1\}$, and $C$ (copy) where each token copies the previous one.
\revised{Under off-policy distillation, the SFT loss of $U$ is given by
\begin{equation}
    \mathbb E_{p,a_1,a_2\sim\mathcal D}[-\log p^U(a_1,a_2|p)]=\mathbb E_{p,a_1,a_2\sim\mathcal D}[-\log(\tfrac{1}{2}\cdot \tfrac{1}{2})]=\log 4.
\end{equation}
\revised{It can be shown that $U$ is the optimal policy under off-policy distillation.}
On the other hand, $C$ results in infinite loss as $p^C(01|0)=p^C(10|1)=0$.}
This creates a pathological situation where off-policy distillation favors $U$ over $C$, despite $C$ perfectly solving both tasks while $U$ solves neither.

In contrast, on-policy distillation optimizes the reverse KL divergence
\begin{equation}
    \mathbb{E}_{p\sim\{0,1\},a_1,a_2\sim p_\text{student}(\cdot|p)}[\log p_\text{student}(a_1,a_2|p)-\log p_\text{expert}(a_1,a_2|p)],
\end{equation}
where $p_\text{expert}$ denotes the \revised{optimal policy under the corresponding domain, i.e., always generating $a_2=p$}.
\revised{$U$ generates $(p,a_1,a_2)=(0,0,1)$ with probability 1/4, to which $p_\text{expert}$ assigns zero probability. This causes $U$ to incur infinite loss under the on-policy distillation objective. On the other hand, in both domains $p=0$ and $p=1$, $C$ generates $(p,a_1,a_2)=(0,0,0)$ and $(p,a_1,a_2)=(1,1,1)$ with probability 1, respectively, which results in a loss of $\frac12\left((\log 1-\log 1/2)+(\log 1-\log 1/2)\right)=\log 2$. It can be observed that on-policy distillation correctly favors $C$ over $U$.}

However, on-policy distillation fails without proper initialization. \revised{Recall that the action space of $a_2$ is a sufficiently large set $\mathcal S$.} If $p_2^M$ is a uniform distribution over the large set $\mathcal{S}$, the student almost never samples tokens 0 or 1. Consequently, the gradient signal only indicates which incorrect tokens to avoid, never providing positive reinforcement for the correct tokens 0 and 1.

\subsection{Proposed Method}
Building on our analysis, we propose a two-phase distillation approach. First, we apply off-policy distillation to establish adequate multi-task performance. Second, we apply on-policy distillation for refinement, which exploits the mode-seeking property of reverse KL divergence. This encourages the model to commit to a subset of modes that fit within the model's limited capacity, rather than attempting to average across multiple modes.

Under our two-phase method, off-policy distillation enables $p_2$ to concentrate probability mass on the correct tokens 0 and 1.
Consequently, leveraging the mode-seeking property of on-policy distillation, the policy may converge to either $C$ (copy policy) or $F$ (flipping policy, where each token is flipped from the previous token), both of which perfectly solve the tasks. Training curves for this toy example are provided in Appendix~\ref{sec:toy_curve}.

For on-policy distillation, we use a batch size of 16 with 16 rollouts per training prompt. We train for 10 steps when combining two tasks (\texttt{airline} and \texttt{telecom}) and 30 steps when combining all four tasks. As shown in Tab.~\ref{tab:main} and Tab.~\ref{tab:main_30b}, the combined approach (``Off. + On.'') successfully recovers performance across all four tasks, matching or exceeding single-task RL baselines.

\subsection{Empirical Validation}\label{sec:validate}
To validate our hypothesis, we conduct additional experiments on tasks \texttt{airline} and \texttt{telecom}.

\paragraph{Collecting rollouts from the multi-task model.} Instead of collecting training data from single-task RL experts, we collect rollouts from the unified multi-task model trained with our two-phase approach (Off. + On.). As shown in Tab.~\ref{tab:modes}, this ``Rollout from Unified'' method achieves performance comparable to single-task RL baselines. While not practically useful (since this requires an existing multi-task model), this experiment validates that the performance degradation stems from aggregating data across multiple expert models, rather than from limitations of the off-policy distillation algorithm itself.

\paragraph{Filtering distillation data.} Since on-policy distillation successfully addresses the performance degradation of off-policy distillation, and our hypothesis attributes this success to its mode-seeking behavior, we test whether explicitly filtering off-policy data to retain only consistent modes yields similar benefits. Specifically, we filter the off-policy dataset at the token level using the ``Off. + On.'' model. For each prompt-answer pair $(x,y)$, we retain token $y_i$ if and only if
\begin{equation}
    \log\pi_\text{Off.+On.}(y_i|xy_{1:i-1}) - \log\pi_\text{RL}(y_i|xy_{1:i-1}) > -0.01,
\end{equation}
where $\pi_\text{RL}$ denotes the corresponding single-task RL model that generated the trajectory. This filtering process removes approximately 18\% of training tokens. We double the amount of distillation data to compensate for the filtered tokens. As presented in Tab.~\ref{tab:modes}, filtering substantially alleviates the performance degradation. The drop on \texttt{telecom} is reduced from -9.9\% to -0.8\% pass\^{}4, supporting our hypothesis that the mode-covering nature of forward KL causes off-policy distillation to average across different modes, and that reducing the number of modes mitigates this issue.

\begin{table}[t]
    \centering
    \small
    \begin{tabular}{c|c|c|c|c}
        \toprule
        \multirow{2}{*}{Method} & \multicolumn{2}{c|}{airline} & \multicolumn{2}{c}{telecom} \\
         & pass@1 & pass\^{}4 & pass@1 & pass\^{}4 \\
        \midrule
        Rollout from Unified & 51.3\deltadown{2.2} & 32.8\deltaneg{-0.8} & 86.0\deltaneg{+0.1} & 68.7\deltaneg{+0.9} \\
        Filtered Data & 51.8\deltadown{1.7} & 31.8\deltadown{1.8} & 83.4\deltadown{2.5} & 67.0\deltaneg{-0.8} \\
        \midrule
        \multicolumn{5}{c}{\textit{Baselines}} \\
        \midrule
        (Multi-Task) Off-Policy Distill. & 52.1\deltadown{1.4} & 31.3\deltadown{2.3} & 79.8\deltabigdown{6.1} & 57.9\deltabigdown{9.9} \\
        Single-Task RL & 53.5 & 33.6 & 85.9 & 67.8 \\
        \bottomrule
    \end{tabular}
    \caption{Ablation studies validating that mode-covering behavior causes performance degradation in multi-task off-policy distillation. ``Rollout from Unified'' collects rollouts from a unified multi-task model (trained with off-policy + on-policy distillation) instead of individual single-task RL experts. ``Filtered Data'' retains only tokens consistent with the unified model when distilling from single-task experts. All methods substantially mitigate the performance degradation observed in naive multi-task off-policy distillation, achieving performance comparable to single-task RL baselines.}
    \label{tab:modes}
\end{table}

\paragraph{Discussions.} Our analysis reveals that the challenges of off-policy distillation stem from the interaction between limited student model capacity and the mode-covering nature of forward KL divergence. A natural question arises: is this a genuine research problem, or will it resolve naturally as models scale? We argue that this remains an important issue for several reasons. First, given computational constraints, post-training of smaller models continues to play a crucial role in LLM research and applications. Second, as models scale, the number of modes in each single-task expert also grows with improved RL performance, maintaining the capacity mismatch. Indeed, our empirical findings in Tab.~\ref{tab:main_30b} demonstrate that this issue persists even in \texttt{30B-A3B} models.
\subsection{Ablation Study on Off-Policy Distillation Data}\label{sec:data_mix}
An important design choice in multi-task off-policy distillation is how to aggregate training data from different tasks. To investigate the impact of data mixture in off-policy distillation, we conduct experiments on \texttt{airline} and \texttt{telecom} with four different configurations: 400:400 (as in our main experiments), 400:800, 400:1200, and 800:800, where $M:N$ denotes $128\times M$ samples from \texttt{airline} and $128\times N$ samples from \texttt{telecom}. For each configuration, we apply an on-policy distillation phase with balanced 1:1 data mixture following off-policy distillation.

The results presented in Tab.~\ref{tab:ratio} reveal the following key insights.
First, simply increasing data volume uniformly does not alleviate performance degradation in multi-task off-policy distillation. The 800:800 mixture exhibits similar performance drops as 400:400.
Second, allocating more data to \texttt{telecom} can partially mitigate its performance drop, though the degradation persists (400:1200 still shows -3.3\% on \texttt{telecom} pass\^{}4). Moreover, this introduces potential trade-offs with other tasks.
Finally, regardless of data mixture, subsequent on-policy distillation consistently addresses the performance degradation issue across all configurations.

\begin{table}[t]
    \centering
    \small

    \vspace{-0.1in}
    \begin{tabular}{c|c|c|c|c}
        \toprule
        \multirow{2}{*}{Ratio} & \multicolumn{2}{c|}{airline} & \multicolumn{2}{c}{telecom} \\
         & pass@1 & pass\^{}4 & pass@1 & pass\^{}4 \\
        \midrule
        \multicolumn{5}{c}{\textit{Off-policy Distill.}} \\
        \midrule
        400 : 400 & 52.1\deltadown{1.4} & 31.3\deltadown{2.3} & 79.8\deltabigdown{6.1} & 57.9\deltabigdown{9.9} \\
        400 : 800 & 49.9\deltabigdown{3.6} & 32.8\deltaneg{-0.8} & 83.6\deltadown{2.3} & 65.3\deltadown{2.5} \\
        400 : 1200 & 50.5\deltabigdown{3.0} & 30.6\deltabigdown{3.0} & 83.7\deltadown{2.2} & 64.5\deltabigdown{3.3} \\
        800 : 800 & 50.9\deltabigdown{2.6} & 31.5\deltadown{2.1} & 79.6\deltabigdown{6.3} & 59.3\deltabigdown{8.5} \\
        \midrule
        \multicolumn{5}{c}{\textit{Off-policy Distill. + On-policy Distill.}} \\
        \midrule
        400 : 400 & 53.1\deltaneg{-0.4} & 33.3\deltaneg{-0.3} & 85.8\deltaneg{-0.1} & 72.1\deltaup{4.1} \\
        400 : 800 & 53.5\deltaneg{+0.0} & 33.1\deltaneg{-0.5} & 86.1\deltaneg{+0.2} & 70.1\deltaup{2.3} \\
        400 : 1200 & 53.1\deltaneg{-0.4} & 35.5\deltaup{1.9} & 86.5\deltaneg{+0.6} & 71.4\deltaup{3.6} \\
        800 : 800 & 52.4\deltadown{1.1} & 33.7\deltaneg{+0.1} & 85.9\deltaneg{+0.0} & 70.5\deltaup{2.7} \\
        \midrule
        \multicolumn{5}{c}{\textit{RL Baseline}} \\
        \midrule
        Single-Task RL & 53.5 & 33.6 & 85.9 & 67.8 \\
        \bottomrule
    \end{tabular}
    \caption{Effect of data mixture on off-policy distillation performance. $M:N$ denotes $128\times M$ samples from \texttt{airline} and $128\times N$ samples from \texttt{telecom} used in off-policy distillation. While off-policy distillation alone suffers from performance degradation regardless of data mixture, a subsequent on-policy distillation phase (always using balanced 1:1 data) consistently recovers performance across all configurations. Colored subscripts indicate performance deltas relative to single-task RL baselines.}
    \label{tab:ratio}
    \vspace{-0.1in}
\end{table}
\section{Conclusion}
This paper investigates distillation-based methods for training multi-task agentic LLMs. Through our experiments, we reveal distinct limitations of off-policy and on-policy distillation approaches. While off-policy distillation is straightforward to implement, its mode-covering behavior makes it problematic in multi-task settings. Conversely, on-policy distillation requires the student model to already possess adequate initial performance. To address these limitations, we propose a two-phase training recipe: first applying off-policy distillation to establish adequate baseline performance, then leveraging on-policy distillation for refinement. We evaluate our method on conversational agents using $\tau^2$-bench and text-based games in GEM, with experiments confirming the effectiveness of our approach. Additionally, we evaluate two alternative strategies: multi-task RL and parameter merging in Appendix~\ref{sec:additional_results}. Under equivalent training budgets, multi-task RL underperforms distillation-based methods. Furthermore, we demonstrate that parameter merging methods exhibit significant performance degradation and are highly sensitive to SFT training dynamics.

\bibliography{colm2026_conference}
\bibliographystyle{colm2026_conference}

\appendix
\section{Training Details}
All training is performed using the AReaL~\citep{fu2025areal} framework.

For $\tau^2$-bench tasks, we follow the method in \citet{gao2026selfevolvingsyntheticdataverifiablereward}. For 8B models, which are not covered in their paper, we adopt a batch size of $16\times 16$ (16 training prompts with 16 rollouts each) and a learning rate of $5\times 10^{-6}$. During training, we use \texttt{Qwen2.5-72B-Instruct} as the user simulator for \texttt{airline} and a fine-tuned \texttt{Qwen3-32B} model as the user simulator for \texttt{telecom}. We use temperature 1 for the user simulator during training and 0 for evaluation.

For GEM tasks, we also use a learning rate of $5\times 10^{-6}$. For \texttt{sudoku}, we adopt a batch size of $16\times 16$. We limit the maximum context length to 16384 during training to reduce training cost. We extend the max number of turns to 60 in evaluation. For \texttt{mastermind}, we adopt a batch size of $256\times 1$. These two tasks do not undergo SFT before RL. Consequently, when conducting on-policy distillation in the multi-task setting, the model is initialized from SFT on only the $\tau^2$-bench tasks.

All tasks have a maximum context length of 32768 during inference. For each turn, the maximum number of generated tokens is 8192.
\section{Additional Results}
\subsection{\revised{Alternative Approaches}}\label{sec:additional_results}
In addition to distillation, we investigate \revised{three} alternative approaches for multi-task learning: (1) off-policy distillation followed by multi-task RL; (2) parameter merging; and \revised{(3) Generalized Knowledge Distillation (GKD)~\citep{agarwal2024onpolicydistillationlanguagemodels}}. For multi-task RL, we apply the same training budget as on-policy distillation (batch size of 16 with 16 rollouts per prompt). Under this setting, we observe that RL training becomes unstable, and we report results from the best checkpoint before training collapse. For parameter merging, we evaluate three methods: naive weight averaging~\citep{wortsman2022modelsoupsaveragingweights}, task arithmetic~\citep{ilharco2023editingmodelstaskarithmetic}, and DARE~\citep{yu2024languagemodelssupermario}. Results are presented in Tab.~\ref{tab:additional}.

\paragraph{Multi-Task RL.} While performance on \texttt{sudoku} and \texttt{mastermind} slightly improves compared to single-task RL baselines, the performance degradation on \texttt{telecom} becomes more severe. This observation aligns with previous findings that optimization across different tasks is imbalanced in the multi-task RL process~\citep{wu2025imbalancedgradientsrlposttraining,ramesh2026multitaskgrporeliablellm}, which presents substantial challenges in practical multi-task settings.

\paragraph{Parameter Merging.} When merging experts from all four tasks, performance drops dramatically except on \texttt{airline}. Analysis reveals that the parameter change $\|\theta_\text{RL}-\theta_\text{pre}\|_2$ for the \texttt{airline} expert is significantly larger than for other tasks, causing \texttt{airline} to dominate the merged model. Notably, this large parameter change primarily originates from the SFT process before RL, demonstrating that parameter merging can be heavily influenced by substantial parameter shifts introduced during SFT. Merging only \texttt{sudoku} and \texttt{mastermind} yields substantially better performance, though significant degradation remains compared to single-task baselines. Interestingly, we observe substantially different behavior when applying parameter merging to math and coding tasks. Using \texttt{DeepSeek-R1-Distill-Qwen-1.5B} with RL training on the boba$^2$ dataset~\citep{fu2025areal}, evaluated on AIME24 and LiveCodeBench v5~\citep{jain2024livecodebenchholisticcontaminationfree}, all three merging methods achieve performance comparable to single-task RL (Tab.~\ref{tab:math_code}), with degradation controlled below 3.0\%. This suggests that parameter merging effectiveness may be task-dependent.

\revised{\paragraph{Generalized Knowledge Distillation.} GKD optimizes the divergence between student and teacher using samples from both the student model and the teacher model. We investigate whether the off-policy distillation phase is necessary, or if single-phase GKD suffices for multi-task distillation. As presented in Tab.~\ref{tab:additional}, single-phase GKD results in significant performance drop on \texttt{telecom}. This suggests that an off-policy distillation phase is indeed required to stabilize on-policy distillation training.}

\begin{table}[t]
    \centering
    \small
    \begin{tabular}{c|l|l|l|l|l|l|l|l}
        \toprule
        \multirow{3}{*}{Method} & \multicolumn{4}{c|}{$\tau^2$-bench} & \multicolumn{4}{c}{GEM} \\
         & \multicolumn{2}{c|}{airline} & \multicolumn{2}{c|}{telecom} & \multicolumn{2}{c|}{sudoku} & \multicolumn{2}{c}{mastermind} \\
         & \multicolumn{1}{c|}{pass@1} & \multicolumn{1}{c|}{pass\^{}4} & \multicolumn{1}{c|}{pass@1} & \multicolumn{1}{c|}{pass\^{}4} & \multicolumn{1}{c|}{pass@1} & \multicolumn{1}{c|}{pass\^{}4} & \multicolumn{1}{c|}{pass@1} & \multicolumn{1}{c}{pass\^{}4} \\
        \midrule
        Avg. & 38.0 & 37.0 & 0.1 & 0.0 & 0.0 & 0.0 & 0.0 & 0.0 \\
         & - & - & - & - & 88.2 & 60.5 & 46.6 & 10.3 \\
        \midrule
        TA & 48.5 & 33.1 & 23.7 & 2.4 & 0.0 & 0.0 & 0.3 & 0.0 \\
         & - & - & - & - & 66.9 & 20.1 & 74.8 & 34.1 \\
        \midrule
        TA + DARE & 48.0 & 32.2 & 24.2 & 3.0 & 0.0 & 0.0 & 0.4 & 0.0 \\
         & - & - & 81.9 & 62.2 & 7.3 & 0.0 & 41.8 & 5.9 \\
         & - & - & - & - & 66.7 & 19.6 & 75.8 & 36.5 \\
        \midrule
        Multi-Task RL & 51.8 & 34.1 & 80.6 & 54.9 & 95.9 & 84.6 & 81.3 & 48.0 \\
        \midrule
        GKD & 51.5 & 33.8 & 81.6 & 61.6 & 92.7 & 74.2 & 78.0 & 41.7 \\
        \bottomrule
    \end{tabular}
    \caption{Performance of alternative multi-task learning approaches. ``Multi-Task RL'' denotes off-policy distillation followed by multi-task RL training. \revised{``GKD'' denotes single-phase Generalized Knowledge Distillation.} ``-'' indicates the corresponding RL expert is excluded from the merge. For example, the second row under ``Avg.'' shows naive weight averaging of only the \texttt{sudoku} and \texttt{mastermind} experts. Parameter merging exhibits significant performance degradation, particularly when \texttt{airline} or \texttt{telecom} experts are included. This occurs because the parameter changes $\|\theta_\text{RL}-\theta_\text{pre}\|_2$ for these tasks substantially exceed those of \texttt{sudoku} and \texttt{mastermind}, causing the corresponding single-task models to dominate the merged model. Multi-task RL demonstrates imbalanced optimization across tasks: while performance on \texttt{sudoku} and \texttt{mastermind} improves, the degradation on \texttt{telecom} becomes more severe than with off-policy distillation alone. \revised{GKD without an off-policy distillation phase also exhibits significant performance drop on \texttt{telecom}.}}
    \label{tab:additional}
\end{table}

\begin{table}[t]
    \centering
    \begin{tabular}{c|c|c}
        \toprule
        Method & AIME24 & LCB v5 \\
        \midrule
        Single-Task RL & 42.8 & 26.4 \\
        \midrule
        Avg. & 40.6 & 23.4 \\
        TA & 41.0 & 25.1 \\
        TA + DARE & 40.8 & 25.6 \\
        \bottomrule
    \end{tabular}
    \caption{Parameter merging performance on math and coding tasks. All three parameter merging methods achieve results comparable to single-task RL baselines, with degradation below 3.0\%.}
    \label{tab:math_code}
\end{table}

\subsection{Toy Example Details}\label{sec:toy_curve}
\begin{figure}[t]
    \centering
    \begin{minipage}[t]{.48\textwidth}    
        \centering
        \includegraphics[width=\textwidth]{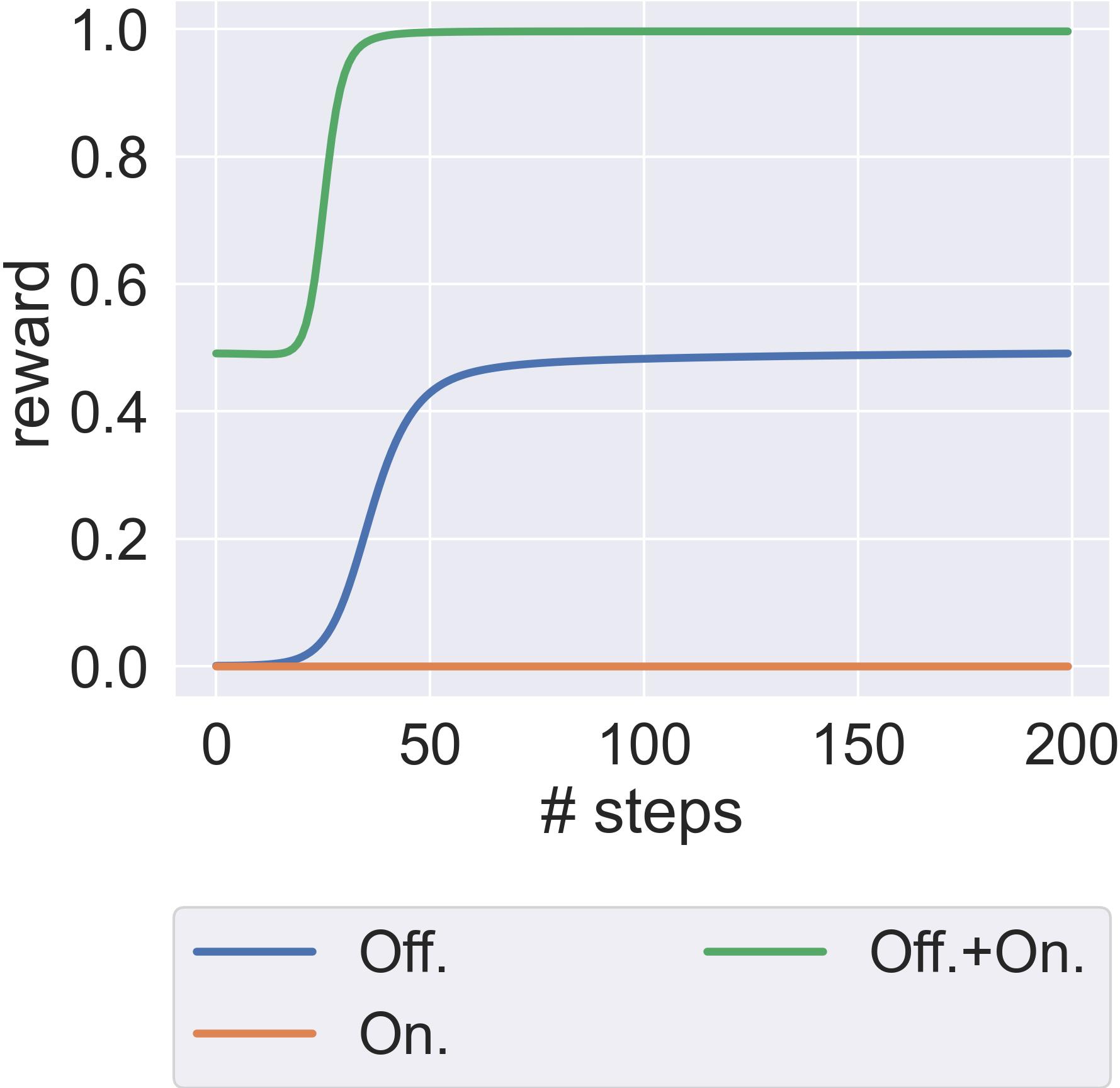}
        \caption{Training curves for our toy example. For ``Off.+On.'', reward is plotted against on-policy distillation steps. Off-policy alone (``Off.'') saturates at reward 0.5 and on-policy alone (``On.'') fails at 0, while our two-phase approach achieves full reward 1.}
        \label{fig:toy_curve}
    \end{minipage}
    \hfill
    \begin{minipage}[t]{.48\textwidth}  
        \centering
        \includegraphics[width=\textwidth]{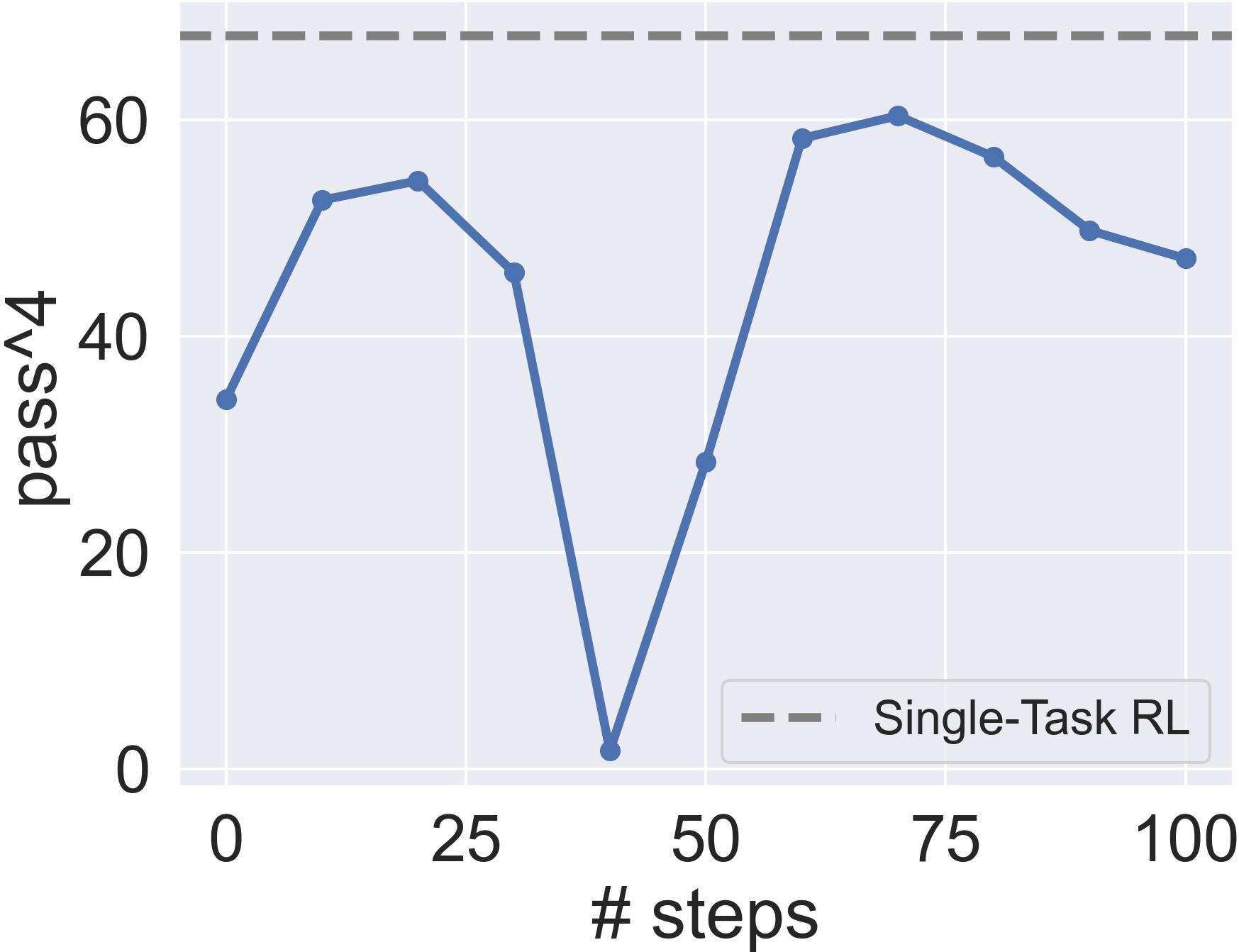}
        \caption{(Multi-Task) On-policy distillation training curves. We plot the \texttt{telecom} pass\^{}4 on 8B models. The model performance stays below single-task RL baseline.}
        \label{fig:onpolicy_curve}
    \end{minipage}
\end{figure}

Fig.~\ref{fig:toy_curve} shows training curves for our toy example (Sec.~\ref{sec:toy}). Off-policy distillation alone saturates at average reward 0.5, while on-policy alone fails at 0. Our two-phase approach achieves full reward 1. We train with gradient descent (learning rate 0.1, momentum 0.9). For both on-policy distillation and the two-phase approach (Off.+On.), we slightly perturb $p_1$ before on-policy distillation to break symmetry: $p_1(a_1=1|0)=\operatorname{sigmoid}(0.02)$ and $p_1(a_1=1|1)=\operatorname{sigmoid}(0.01)$. This perturbation reflects realistic conditions where models naturally produce different distributions across tasks.

\subsection{On-Policy Distillation Training Curves}
Fig.~\ref{fig:onpolicy_curve} shows on-policy distillation training curves for multi-task learning (all 4 tasks), plotting \texttt{telecom} pass\^{}4 on 8B models. Without proper initialization, performance remains below the single-task RL baseline even after 100 steps. In contrast, our two-phase approach recovers baseline performance in 30 steps.

\end{document}